\pdfoutput=1

\documentclass[11pt]{article}

\usepackage{emnlp2021}

\usepackage{times}
\usepackage{latexsym}

\usepackage[T1]{fontenc}

\usepackage[utf8]{inputenc}

\usepackage{microtype}

\title{Euphemistic Phrase Detection by Masked Language Model}

\author{Wanzheng Zhu \and Suma Bhat\\
	University of Illinois at Urbana-Champaign, USA \\
	\texttt{wz6@illinois.edu}, \texttt{spbhat2@illinois.edu}}

\usepackage{graphicx, amsmath, amsfonts, amssymb, xspace, color, enumitem}
\usepackage{amsthm}
\usepackage{amsfonts}
\usepackage{xcolor}
\usepackage{booktabs}
\usepackage{tabularx} 
\usepackage[noend,ruled,linesnumbered]{algorithm2e}
\usepackage{multirow}
\usepackage{subcaption}

\newcommand{\etc}{{etc.}\xspace} 
\newcommand{\ie}{{i.e.}\xspace} 
\newcommand{\eg}{{e.g.}\xspace} 
\newcommand{\our}{\textsc{EPD}\xspace}
\newcommand{\nop}[1]{}

\newcommand{\pb}[1]{\textbf{\textcolor{violet}{#1}}}

\usepackage{comment}

\begin{document}
\maketitle
\begin{abstract}
	It is a well-known approach for fringe groups and organizations to use \emph{euphemisms}---ordinary-sounding and innocent-looking words with a secret meaning---to conceal what they are discussing. 
	For instance, drug dealers often use ``pot'' for marijuana and ``avocado'' for heroin. 
	From a social media content moderation perspective, though  recent advances in NLP have enabled the automatic detection of such \textit{single-word} euphemisms, no existing work is capable of automatically detecting \textit{multi-word} euphemisms, such as ``blue dream'' (marijuana) and ``black tar'' (heroin). 
	Our paper tackles the problem of \textbf{euphemistic phrase detection} without human effort for the first time, as far as we are aware. 
	We first perform phrase mining on a raw text corpus (\eg, social media posts) to extract quality phrases. 
	Then, we utilize word embedding similarities to select a set of euphemistic phrase candidates. 
	Finally, we rank those candidates by a masked language model---SpanBERT. 
	Compared to strong baselines, we report 20-50\% higher detection accuracies using our algorithm for detecting euphemistic phrases.
\end{abstract}

\section{Introduction}
\label{sec:introduction}
Euphemisms---ordinary-sounding and innocent-looking words---have long been used in human communication as an instrument to conceal secret information \cite{bellman1981paradox}. A primary motive of their use on social media is to evade automatic content moderation efforts enforced by such platforms \cite{Ofcom:AI2019,yuan2018reading}.
For example, a rich lexicon of drug euphemisms has evolved over time, with entire communities subscribing to benign sounding words that allude to drug names (\eg, \{``popcorn'', ``blueberry'', ``green crack'', ``blue dream''\} $\longrightarrow$ ``marijuana'', \{``coke'', ``white horse'', ``happy powder''\} $\longrightarrow$ ``cocaine'').

Research on automatic euphemism detection has recently received increased attention in the natural language processing communities 
\cite{durrett2017identifying,magu2018determining,pei2019slang,felt2020recognizing}, and the security and privacy communities \cite{zhao2016chinese,yang2017learn,yuan2018reading,takuro2020codewords,zhu2021selfsupervised}. 
However, existing approaches can only detect \textit{single-word} euphemisms (\eg, ``popcorn'', ``coke''), and fail to detect \textit{multi-word} euphemisms (\eg, ``black tar'', ``cbd oil'') automatically. 
Therefore, offenders can simply invent euphemistic phrases to evade  content moderation and thwart censorship.

Our paper focuses on the task of \textbf{euphemistic phrase detection}---detecting phrases that are used as euphemisms for a list of target keywords---by extending the state-of-the-art single-word euphemism detection algorithm proposed by \citet{zhu2021selfsupervised}. 
Our proposed approach first mines quality phrases from the text corpus using AutoPhrase \cite{shang2018automated,liu2015mining}, a data-driven phrase mining tool. 
Then, it filters noisy candidates that are not semantically related to any of the target keywords  (\eg, heroin, marijuana in the drug category). 
This serves as a pre-selection step to construct a euphemistic phrase candidate pool. 
Finally, we rank the pre-selected candidates using SpanBERT \cite{joshi2020spanbert}, a pre-training Masked Language Model (MLM) that is designed to better predict the span of tokens (\ie, phrases) in text. 

Evaluating on the benchmark drug dataset in \citet{zhu2021selfsupervised}, we find that our proposed approach yields euphemistic phrase detection results that are 20-50\% higher than a set of strong baseline methods. 
A qualitative analysis reveals that our approach also discovers correct euphemisms that were not on our ground truth list, \ie, it can detect previously unknown euphemisms and even new types of drugs. 
This is of significant utility in the context of Internet communities, where euphemisms evolve rapidly and new types of drugs may be invented.

\section{Proposed Model}
\label{sec:model}
In this study, we assume access to a raw text corpus (\eg, a set of posts from an online forum). 
In practice, forum users may use \emph{euphemisms}---words that are used as substitutes for one of the target keywords (\eg, heroin, marijuana). 
We aim to learn which multi-word \textit{phrases} are being used as euphemisms for the target keywords. 
The euphemism detection task takes as input 
(1) the raw text corpus and 
(2) a list of target keywords. 
The output is an ordered ranked list of euphemistic phrase candidates, sorted by model confidence. 

\begin{figure}[t]
	\centering
	\includegraphics[width=0.98\linewidth]{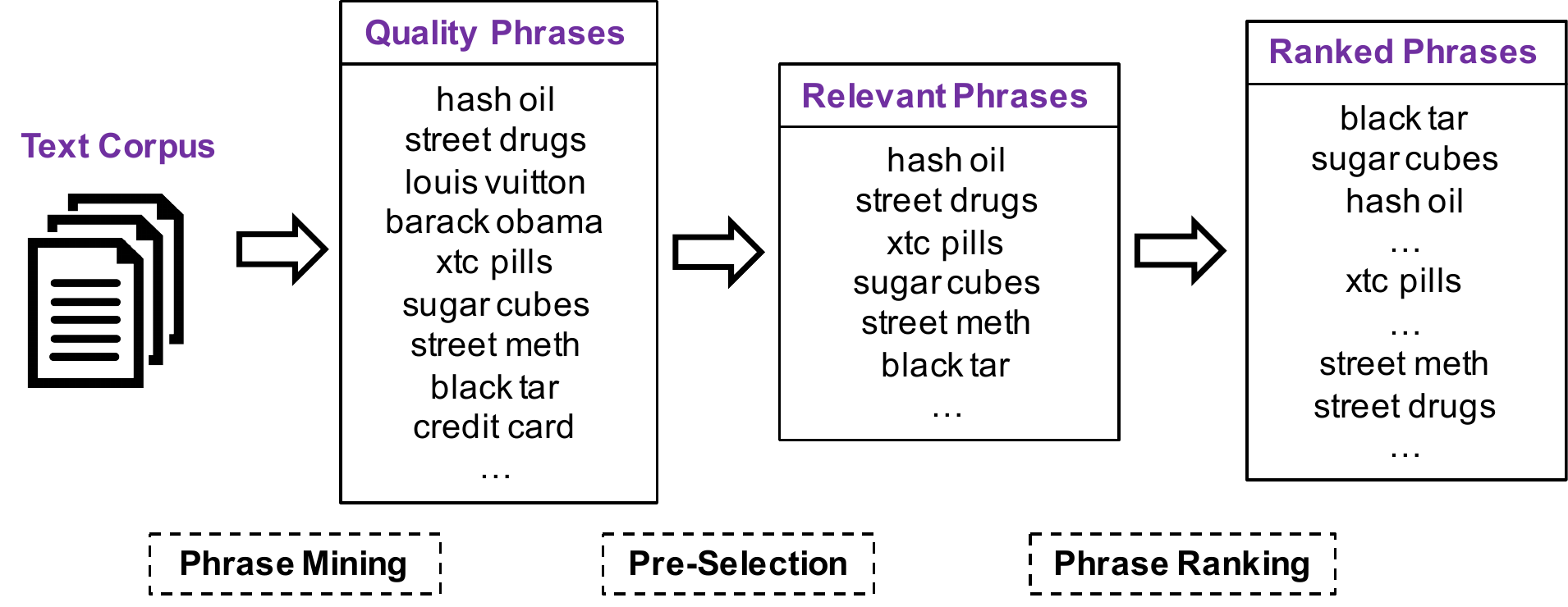}
	\caption{An overview of our proposed framework}
	\label{fig:model}
\end{figure}

Our proposed approach for euphemistic phrase detection has three stages (shown in Figure \ref{fig:model}): 
1) Mining quality phrases, 
2) Pre-selecting euphemistic phrase candidates using cosine similarities of word2vec embeddings \cite{mikolov2013efficient,mikolov2013distributed}, and 
3) Ranking euphemistic phrases with a masked language model.

\subsection{Quality Phrase Mining}
\label{sec:model_phrase}
Phrase mining aims to generate a list of quality phrases, which serves as the candidate pool for the algorithm to rank. 
We select AutoPhrase \cite{shang2018automated,liu2015mining}, which has demonstrated superior phrase mining performance in a wide range of settings, to mine quality phrases. This is because we are interested in a data-driven method of detection from a domain-specific text corpus such as subreddit\footnote{Forums hosted on the Reddit website, and associated with a specific topic.}, rather than by using trained linguistic analyzers (\eg, dependency parsers) that are less likely to have a satisfactory performance on text corpora with unusual usage of words (euphemisms). 
By incorporating distant supervision (\ie, Wikipedia) and part-of-speech tags as \citet{shang2018automated}, we empirically find that AutoPhrase can extract meaningful phrases successfully.

\subsection{Pre-Selection of Phrase Candidates}
\label{sec:model_pre}
AutoPhrase takes only a text corpus as its input and  produces phrases that may or may not be relevant  to any of the target keywords. 
This stage aims to filter out  phrases that are not relevant to the target keywords and thus pre-select the euphemistic phrase candidates. 
This serves to not only pre-filter noisy candidates, but also to reduce the computational resources in the subsequent ranking algorithm. 

Specifically, we use the word2vec algorithm \cite{mikolov2013efficient,mikolov2013distributed} to learn the embeddings for all the words and phrases.\footnote{We use the Gensim package in Python3 for word2vec training. We use a context window of 6, an embedding dimension of 100, a minimum count of 5, and a sampling rate of $10^{-4}$.} 
Relying on the distributional hypothesis that semantically similar words occur in linguistically similar contexts, we assume that the euphemistic phrases should not be too far from the target keywords on the embedding space. 
Therefore, we select the top $k$ phrases\footnote{We empirically set $k=1000$ in our experiments.} in terms of the cosine similarities between the embeddings of each extracted phrase and the average embeddings of all target keywords. 

\subsection{Euphemistic Phrase Ranking}
\label{sec:model_rank}
We extract contextual information of the target keywords and filter out uninformative contexts, following \citet{zhu2021selfsupervised}. 
Next, with a collection of informative masked sentences (\eg, ``This 22 year old former [MASK] addict who I did drugs with was caught this night''), we aim to rank the pre-selected phrase candidates for their ability to serve as a replacement of the masked keyword. 
Toward ranking the candidates for filling in the mask, a common approach is to use BERT \cite{devlin2019bert}, but BERT can be used to only rank single words. 
Here, we leverage the idea of masked language model applied not at the word level, but at the phrase level to facilitate detection.
Therefore, we select SpanBERT \cite{joshi2020spanbert} to rank the candidates, because it is designed to better represent and predict contiguous spans of text and it enables the likelihood calculation of multi-word candidates in a given context. 

We fine-tune the pre-trained SpanBERT model with the text corpus of interest.\footnote{\url{https://github.com/facebookresearch/SpanBERT}.} 
Then, for each masked sentence $m$, and for each phrase candidate $c$, we compute its MLM probability (the probability of the phrase $c$ occurring in $m$ as predicted by the masked language model) $h_{c,m}$  by the fine-tuned SpanBERT model. 
Therefore, given a set of masked sentences, the weight $w_{c}$ of a word candidate $c$ is calculated as: 
$w_c = \sum_{m'}h_{c, m'}$. 
Lastly, we  rank the phrase candidates by their weights.

\section{Empirical Evaluation}
\label{sec:exp}
We evaluate our proposed model (denoted as ``EPD'') and the following baselines on the benchmark \textit{drug} dataset in \citet{zhu2021selfsupervised}, and compare it with the following baseline models:
\begin{itemize}[leftmargin=*]
	\item \textbf{SentEuph} \cite{felt2020recognizing} recognizes euphemisms by sentiment analysis and a bootstrapping algorithm for semantic lexicon induction. 
	For a fair comparison, we do not include its manual filtering stage and exclude the single-word predictions from the output. 
	\item \textbf{Word2vec}: we follow Section \ref{sec:model_phrase} and \ref{sec:model_pre} to rank all phrases by the cosine similarities between each phrase and the input target keywords. We do not include the final euphemistic phrase ranking step in Section \ref{sec:model_rank}.
	This is one of the most straightforward baselines and also, an ablation study to investigate the effectiveness of the euphemistic phrase ranking step. 
	\item \textbf{EigenEuph} \cite{magu2018determining} leverages word and phrase embeddings (following Section \ref{sec:model_phrase} and \ref{sec:model_pre}) and a community detection algorithm, to generate a cluster of euphemisms by the ranking metric of eigenvector centralities. 
	\item \textbf{\our-rank-all} is a simpler version of \our. It does not pre-select euphemistic phrase candidates described in Section \ref{sec:model_pre} but uses SpanBERT to rank \textit{all} phrases mined by AutoPhrase. 
	\item \textbf{\our-ILM} ranks the pre-selected phrase candidates by Infilling by Language Modeling (ILM)\footnote{\url{https://github.com/chrisdonahue/ilm}} 
	\cite{donahue2020enabling} instead of SpanBERT. ILM is optimized for predicting fixed-length missing tokens of a document. We set the token length to be 2, since a majority of euphemistic phrases (\ie, 749 out of 820 in the drug dataset) have 2 words. 
\end{itemize}
Following \citet{zhu2021selfsupervised}, we use the evaluation metric \textbf{precision at k} ($P@k$) to compare the generated candidates of each method with the ground truth list of euphemistic phrases. 
For a fair comparison of the baselines, we experiment with different combinations of parameters and report the best performance for each baseline method.

\begin{table}[t]
	\centering
	\small
	\begin{tabular}{c|cccccccc}
		\toprule
		\multicolumn{1}{c}{} & \textbf{$P@10$} & \textbf{$P@20$} & \textbf{$P@30$} & \textbf{$P@50$} \\
		\midrule
		\textbf{SentEuph} & 0.00 & 0.00 & 0.03  & 0.02 \\
		\textbf{Word2vec} & 0.10 & 0.10 & 0.07 & 0.06 \\
		\textbf{EigenEuph} & 0.10 & 0.15 & 0.13 & 0.10 \\
		\textbf{\our-rank-all} & 0.20 & 0.25 & 0.20 & 0.16 \\
		\textbf{\our-ILM} & 0.00 & 0.10 & 0.10 &  0.12 \\
		\textbf{\our} & \textbf{0.30} & \textbf{0.30} & \textbf{0.27} & \textbf{0.22} \\
		\bottomrule
	\end{tabular}
	\caption{Results on euphemistic phrase detection. Best results are in bold.}
	\label{table:res_dec}
\end{table}

\begin{table}[t]
	\centering
	\small
	\begin{tabular}{p{0.45\textwidth}}
		\toprule
		\multicolumn{1}{c}{\textbf{Euphemistic Phrase Candidates}}\\
		\midrule
		\pb{black tar}, nitric oxide, nitrous oxide, \pb{hash oil}, citric acid, crystal meth, lysergic acid, hydrochloric acid, \pb{cbd oil}, magic mushroom, 
		sour diesel, xtc pills, crystal meth, isopropyl alcohol, \pb{sugar cubes}, speed paste, \pb{og kush}, fentanyl powder, \pb{brown sugar}, pot brownies, 
		xanax bars, hemp oil, \pb{coca cola}, dnm coke, co2 oil, \pb{blue dream}, gold bullion, cannabis tincture, oxy pills, amphetamine powder
		\\
		\bottomrule
	\end{tabular}
	\caption{Top 30 output by \our. Purple bold words are correct detections as marked by the ground truth list.}
	\label{table:output}
\end{table}

\begin{table*}[ht]
	\centering
	\includegraphics[width=0.98\linewidth]{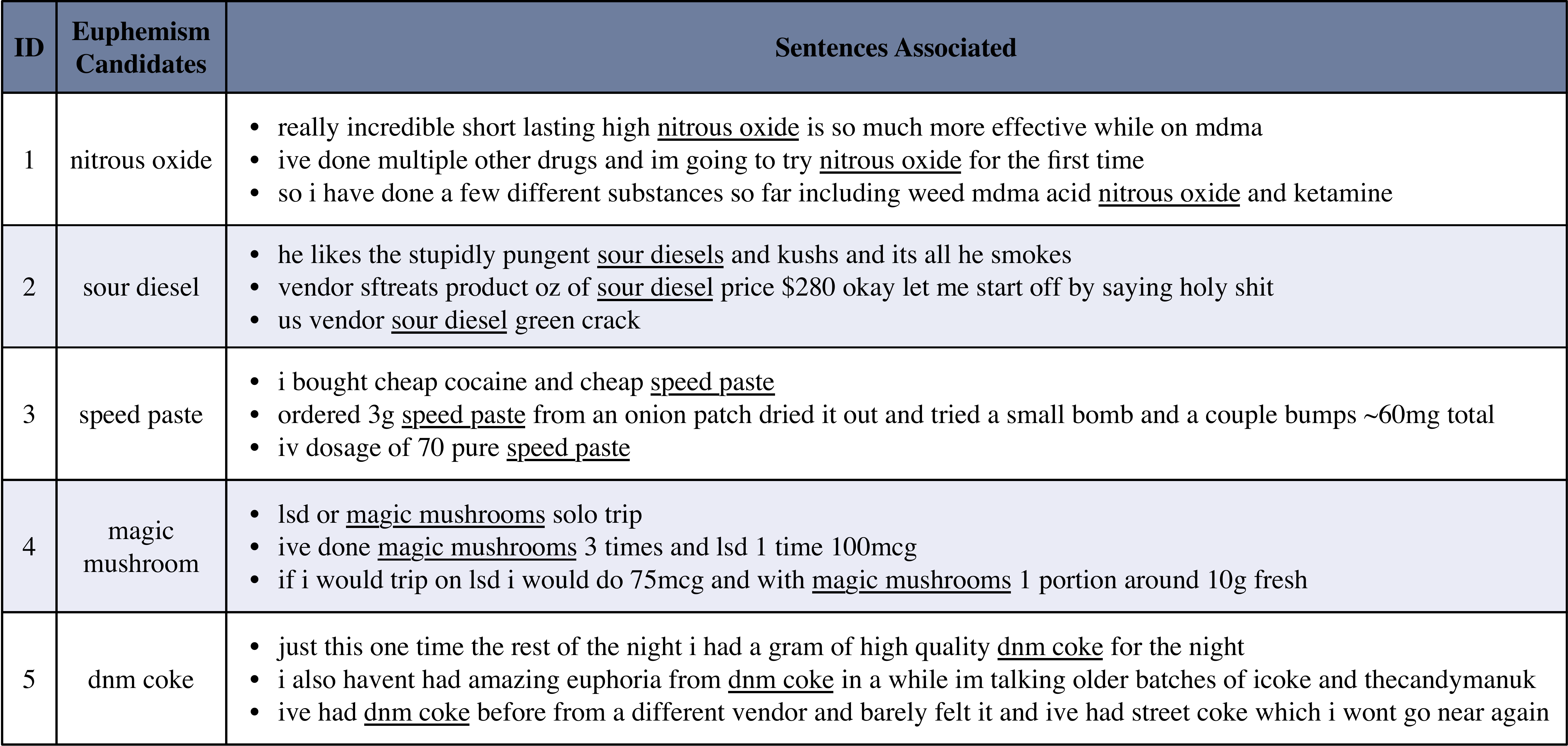}
	\caption{Case Studies of the false positives detected on the drug dataset. They are real examples from Reddit.}
	\label{fig:casestudies-detection}
\end{table*}

\subsection{Results}
Table \ref{table:res_dec} summarizes the euphemistic phrase detection results. 
We note that our proposed approach outperforms all the baselines by a wide margin for the different settings of the evaluation metric. 

SentEuph's poor performance could be attributed to the absence of the required additional manual filtering stage to refine the results. As mentioned before, this was done to compare the approaches based on their automatic performance alone. 

Word2vec is one of the most straightforward baselines. By taking advantage of the distributional hypothesis, it can output some reasonable results. 
However, its performance is still inferior largely because it learns a single embedding for each token and therefore does not distinguish different senses of the same token. 
EigenEuph, which leverages a community detection algorithm to enhance the similarity for different tokens, has slightly better results than the vanilla Word2vec baseline.

By comparing the performance of \our and Word2vec, we conclude that it is effective to adopt SpanBERT for the final ranking of the pre-selected euphemistic phrase candidates. 
Comparing the performance of \our and \our-rank-all, we demonstrate that it is effective to pre-select a set of euphemistic phrase candidates using word2vec before ranking by SpanBERT.\footnote{We also point out that the pre-selection step saves 62\% of the run time in our experiment.}


ILM performs poorly for this task. 
ILM is designed for text infilling for a \textit{document}, but not for a \textit{sentence}. 
By inspecting the output of ILM, we find that many top ranked candidates contain a punctuation which separates one sentence from another. 
For instance, in the masked sentence ``these products can sometimes be found in shitty and dangerous [MASK] [MASK] pills'', ILM ranks "places ." as the best candidates to replace the masks. 
Though we limit the ranking candidates to be the pre-selected phrases generated in Section~\ref{sec:model_pre}, we still find its ranking performance to be suboptimal. 
However, we do find that ILM produces reasonable results for single-word prediction, which is not the task we consider.

\subsection{False Positive Analysis}
We present the top 30 outputs generated by \our in Table~\ref{table:output} and perform case study on the false positives in Table~\ref{fig:casestudies-detection}. 
A closer analysis of the false positives reveals that some of them are true euphemistic phrases for drugs that were not present in the ground truth list 
(\ie, cases 2-5 in Table~\ref{fig:casestudies-detection}). 
This is of significant utility in the context of Internet communities, where memes and slangs lead to rapidly evolving euphemistic vocabulary and new types of drugs may be invented. 
For instance, we discover ``nitrous oxide'' (commonly known as ``laughing gas'', popular among young people). 
Among other false positives, we find that many of them are strongly related to a drug, but they are not proper euphemisms such as ``crystal meth'' and ``xtc pills" (``ecstasy pills'').

\subsection{Generalizability to Other Datasets}
Owing the limited availability or the  nature of euphemisms in the dataset, we perform experiments on only one real-life dataset. 
We did not perform experiments on the weapon and the sexuality datasets used in \citet{zhu2021selfsupervised}, because most euphemisms used are single words rather than multi-word phrases. 
Neither did we perform experiments on the hate speech dataset collected by \citet{magu2018determining} since the dataset was not publicly available.

Despite the lack of empirical support, we believe our approach to be generalizable to other datasets or domains since the algorithm does not make any domain-specific assumptions. 
Besides, \our shares a similar model architecture with the algorithm proposed by \citet{zhu2021selfsupervised}, shown to be robust across various datasets. 
However, we do admit that the generalizability of our approach needs to be justified empirically on multiple real-life datasets. 
We leave the dataset collection and empirical evaluation for future work.

\section{Related Work}
\label{sec:related_work}
Euphemism detection and its related work has recently received increased attention from the natural language processing and security and privacy communities \cite{durrett2017identifying,portnoff2017tools,magu2018determining,pei2019slang,felt2020recognizing,zhao2016chinese,yang2017learn,zhu2019fuse,yuan2018reading,takuro2020codewords,zhu2021selfsupervised}. 
Existing euphemism detection work have established a number of models by supervised \cite{pei2019slang}, semi-supervised \cite{durrett2017identifying} and unsupervised learning schemes \cite{zhao2016chinese,magu2018determining}, on diverse categories and platforms \cite{yang2017learn,takuro2020codewords}, with and without distant-supervision \cite{portnoff2017tools,felt2020recognizing}. 

Without requiring any online search services, one major line of existing work have relied on static word embeddings (\eg, word2vec) in combination with network analysis \cite{taylor2017surfacing,magu2018determining}, sentiment analysis \cite{felt2020recognizing}, and semantic comparison across corpora \cite{yuan2018reading}. 
However, the use of static word embeddings provides a single representation for a given word without accounting for its polysemy, and yields limited benefits. 
Therefore, \citet{zhu2021selfsupervised} propose to explicitly harness the contextual information, formulate the problem as an unsupervised fill-in-the-mask problem \cite{devlin2019bert,donahue2020enabling}, and solve it by a masked language model with state-of-the-art results.

Though prior studies report excellent results, to the best of our knowledge, none of the available approaches is capable of detecting euphemistic phrases without human effort.\footnote{\citet{felt2020recognizing} achieves euphemistic phrases detection, \textit{with} additional manual filtering process.} 
Therefore, policy evaders could simply invent euphemistic phrases to escape from the censorship. 
Our work bridges this gap by extending the state-of-the-art euphemism detection approach proposed by  \citet{zhu2021selfsupervised} and achieves holistic euphemism detection by enabling the detection of euphemistic phrases. 


\section{Conclusion}
\label{sec:conclusion}
We have proposed a solution to address the problem of euphemistic phrase detection. 
By mining quality phrases from the text corpus, pre-selecting euphemistic phrase candidates, and ranking phrases by a masked language model, we, for the first time, achieve euphemistic phrase detection automatically.\footnote{Our code is publicly available at \url{https://github.com/WanzhengZhu/Euphemism}.} 
Moreover, we discover new euphemisms that are not even on the ground truth list, which is valuable for content moderation on social media platforms.

\section*{Acknowledgements}
We thank the anonymous reviewers for their helpful comments on earlier drafts that significantly helped improve this manuscript. 
This research was supported by the National Science Foundation award CNS-1720268.

\section*{Ethical Considerations}
The data we use in this paper are from the previous years, were posted on publicly accessible websites, and do not contain any personal identifiable information (\ie, no real names, email addresses, IP addresses, \etc). 
Just like \citet{zhu2021selfsupervised}, our analyses relying on user-generated content do not constitute human subjects research, and are thus not within the purview of the IRB.\footnote{Readers are referred to the \textit{Ethics} Section in \citet{zhu2021selfsupervised} for more detailed information.} 

\bibliography{ref}
\bibliographystyle{acl_natbib}
\end{document}